\def\year{2022}\relax
\documentclass[letterpaper]{article} 
\usepackage{aaai22}  
\usepackage{times}  
\usepackage{helvet}  
\usepackage{courier}  
\usepackage[hyphens]{url}  
\usepackage{graphicx} 
\usepackage{bbm}
\usepackage{natbib}  
\usepackage{caption} 
\DeclareCaptionStyle{ruled}{labelfont=normalfont,labelsep=colon,strut=off} 
\usepackage{algorithm}
\usepackage{algorithmic}
\usepackage{bbding}
\usepackage{soul,color}

\usepackage{multirow}
\usepackage{booktabs}
\usepackage{amsmath}
\usepackage{calc}

\usepackage{twoopt}
\usepackage{tikz}
\usetikzlibrary{fit}

\usepackage{newfloat}
\usepackage{listings}
\floatstyle{ruled}
\newfloat{listing}{tb}{lst}{}
\floatname{listing}{Listing}
\title{CMUA-Watermark: A Cross-Model Universal Adversarial Watermark for Combating Deepfakes}
\author{
     Hao Huang,\textsuperscript{\rm 1,2}
     Yongtao Wang,\textsuperscript{\rm 1,2}\thanks{Corresponding author}
     Zhaoyu Chen,\textsuperscript{\rm 3}
     Yuze Zhang,\textsuperscript{\rm 1}
     Yuheng Li,\textsuperscript{\rm 1}
     Zhi Tang,\textsuperscript{\rm 1,2} \\
     Wei Chu,\textsuperscript{\rm 4}
     Jingdong Chen,\textsuperscript{\rm 4}
     Weisi Lin,\textsuperscript{\rm 5}
     Kai-Kuang Ma\textsuperscript{\rm 5}
}
\affiliations{
    \textsuperscript{\rm 1}Peking University 
    \textsuperscript{\rm 2}State Key Laboratory of Media Convergence Production Technology and Systems \\
    \textsuperscript{\rm 3}Fudan University
    \textsuperscript{\rm 4}Ant Group
    \textsuperscript{\rm 5}Nanyang Technological University\\
    
    huanghao@stu.pku.edu.cn, \{wyt, kevinzyz, tangzhi\}@pku.edu.cn, zhaoyuchen20@fudan.edu.cn, \\
    yuhengli2021@outlook.com, \{weichu.cw,  jingdongchen.cjd\}@antgroup.com, \{wslin, ekkma\}@ntu.edu.sg

}

\usepackage{bibentry}

\begin{document}

\maketitle

\begin{abstract}

Malicious applications of deepfakes (i.e., technologies generating target facial attributes or entire faces from facial images) have posed a huge threat to individuals' reputation and security. To mitigate these threats, recent studies have proposed adversarial watermarks to combat deepfake models, leading them to generate distorted outputs. Despite achieving impressive results, these adversarial watermarks have low image-level and model-level transferability, meaning that they can protect only one facial image from one specific deepfake model. To address these issues, we propose a novel solution that can generate a Cross-Model Universal Adversarial Watermark (CMUA-Watermark), protecting a large number of facial images from multiple deepfake models. Specifically, we begin by proposing a cross-model universal attack pipeline that attacks multiple deepfake models iteratively. Then, we design a two-level perturbation fusion strategy to alleviate the conflict between the adversarial watermarks generated by different facial images and models. Moreover, we address the key problem in cross-model optimization with a heuristic approach to automatically find the suitable attack step sizes for different models, further weakening the model-level conflict. Finally, we introduce a more reasonable and comprehensive evaluation method to fully test the proposed method and compare it with existing ones. Extensive experimental results demonstrate that the proposed CMUA-Watermark can effectively distort the fake facial images generated by multiple deepfake models while achieving a better performance than existing methods. Our code is available at https://github.com/VDIGPKU/CMUA-Watermark.

\end{abstract}

\section{Introduction}

\begin{figure}[h]
  \centering
  \includegraphics[width=8.5cm]{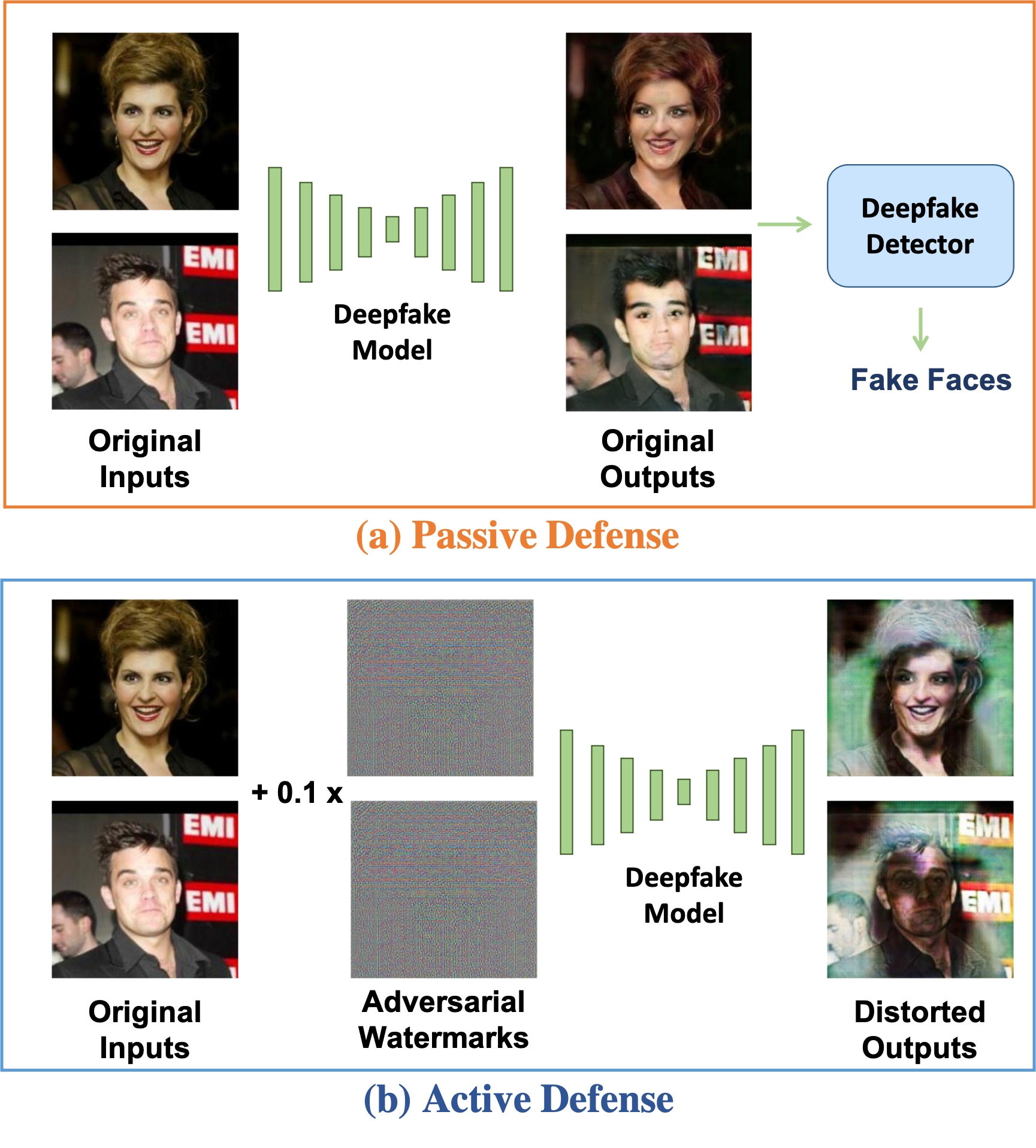}
  \caption{Passive Defense using a deepfake detector can only posteriorly lessen deepfake's harms by detecting the modified images, while active defense uses adversarial watermarks to disrupt the deepfake model to generate perceptibly distorted outputs, thus mitigating risk in advance.}
  \label{fig:method_compare}
\end{figure}

\begin{figure*}[h]
  \centering
  \includegraphics[width=17cm]{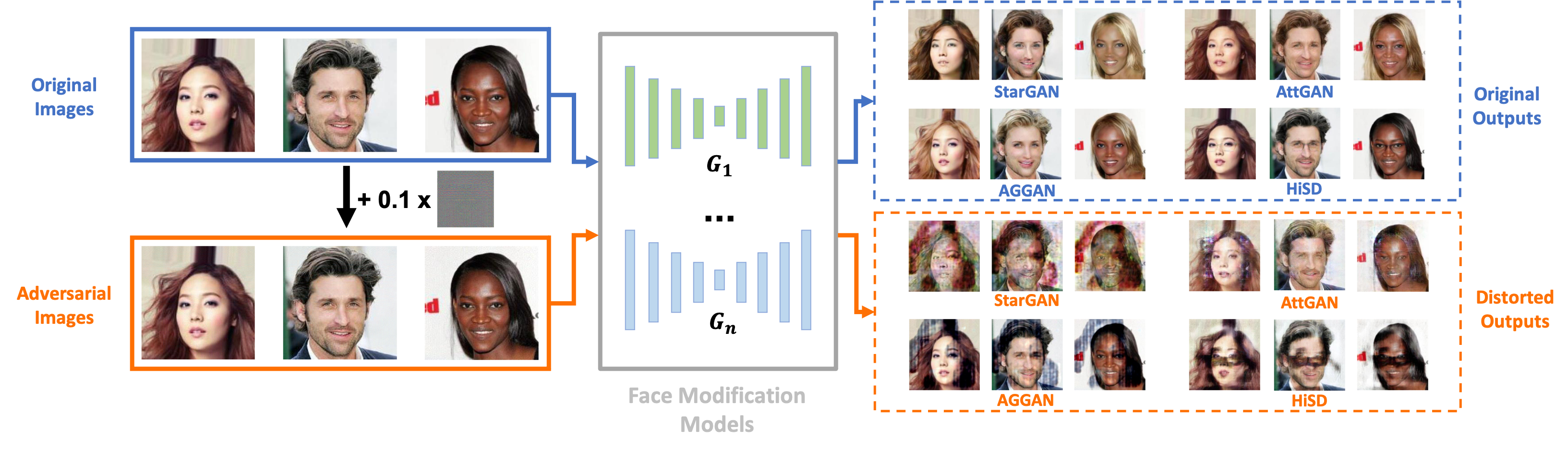}
  \caption{Illustration of our CMUA-Watermark. Once the CMUA-watermark has been generated, we can add it directly to any facial image to generate a protected image that is visually identical to the original image but can distort outputs of deepfake models such as StarGAN \cite{choi2018stargan}, AGGAN \cite{AGGAN}, AttGAN \cite{AttGAN} and HiSD \cite{Li_2021_CVPR}.}
  \label{fig:demo}
\end{figure*}

Recently, improvements of Generative Adversarial Networks (GANs) have shown impressive results in virtual content generation, creating considerable economic and entertainment value. However, deepfakes, deep learning based face modification networks using GANs to generate fake content of target persons or target attributes, have caused great harm to people's privacy and reputation. On one hand, fake images and videos can show things that a person has never said or done, causing damage to his or her reputation, especially when involving pornography or politics \cite{TOLOSANA2020131}. On the other hand, fake facial images with target attributes may pass the biometric authentication of commercial applications, potentially breaching security \cite{DBLP:journals/corr/abs-1812-08685}. Hence, defending threats brought by deepfakes requires not only distorting the modified images and lowering their visual quality to aid humans in distinguishing them from realistic images, but also ensuring that counterfeit faces do not pass liveness detection, which is the first step of most biometric verifications.

Generally speaking, the mainstream way to mitigate the risk of deepfakes is passive defense, i.e., training deepfake detectors to detect modified content \cite{8630761,tariq2021web, zhao2021multiattentional,Sun2021DomainGF,Chen2021LocalRL}. Such detectors are essentially binary classifiers, predicting whether or not an image is fabricated by deepfake models. However, protecting facial images in this way is analogous to closing the stable door after the horse has bolted; the effect and harm caused cannot be reversed entirely and the risk still remains. Recently, \cite{ruiz2020disrupting} suggests using adversarial watermarks to combat deepfake models, leading them to generate visibly unreal outputs. Since these watermarks can be added to facial images in advance, they can avert the risk of malicious deepfake usage afterward as an active defense. The schematic comparison of the two ways is illustrated in Figure \ref{fig:method_compare}. Though \cite{ruiz2020disrupting} could defend potential threats, it can only generate an image-and-model-specific adversarial watermark, meaning that the watermark can only protect one facial image against one specific deepfake model.

To address these issues, we propose an effective and efficient solution in this work, which uses a small number (as small as 128) of training facial images to craft a cross-model universal adversarial watermark (CMUA-Watermark) for protecting a large number of facial images from multiple deepfake models, as depicted in Figure \ref{fig:demo}. Firstly, we propose a cross-model universal attack approach based on a vanilla attack method that can only protect one specific image from one model, i.e., PGD \cite{madry2018towards}. Specifically, to alleviate the conflict among adversarial watermarks generated from different images and models, we newly propose a two-level perturbation fusion strategy (i.e., image-level fusion and model-level fusion) during the cross-model universal attack process. Secondly, to further weaken the conflict among adversarial watermarks generated from different models thus enhancing the transferability of the generated CMUA-Watermark, we exploit the Tree-Structured Parzen Estimator (TPE) \cite{NIPS2011_86e8f7ab} algorithm to automatically find the attack step sizes for different models.

Moreover, the existing evaluation method in \cite{ruiz2020disrupting} is not reasonable and comprehensive enough. Firstly, measuring image distortion through directly calculating $L^1$ or $L^2$ distances between entire original and distorted outputs overlooks deepfakes that only modify a few attributes (e.g., HiSD \cite{Li_2021_CVPR} can add only a pair of glasses), as the measured distortion will be averaged by other unchanged areas. Instead, we propose using a modification mask to focus more on the modified areas. Secondly, solely considering $L^1$ or $L^2$ distances is not sufficient; ensuring protection also requires metrics reflecting generation quality and biological characteristics of distorted outputs. Accordingly, we use Fréchet Inception Distance (FID) \cite{FID} to measure generation quality and exploit the confidence scores as well as the passing rates of liveness
detection models to measure the biological characteristics of distorted outputs.

Our contributions can be summarized as the following:

\begin{itemize}

\item We are the first to introduce the novel idea of generating a cross-model universal adversarial watermark (CMUA-Watermark) to protect human facial images from multiple deepfakes, needing only 128 training facial images to protect a myriad of facial images.

\item We propose a simple yet effective perturbation fusion strategy to alleviate the conflict and enhance the image-level and model-level transferability of the proposed CMUA-watermark.

\item We deeply analyze the cross-model optimization process and develop an automatic step size tuning algorithm to find suitable attack step sizes for different models.

\item We introduce a more reasonable and comprehensive evaluation method to fully evaluate the effectiveness of the adversarial watermark on combating deepfakes.

\end{itemize}

\begin{figure*}[h]
  \centering
  \includegraphics[width=17cm]{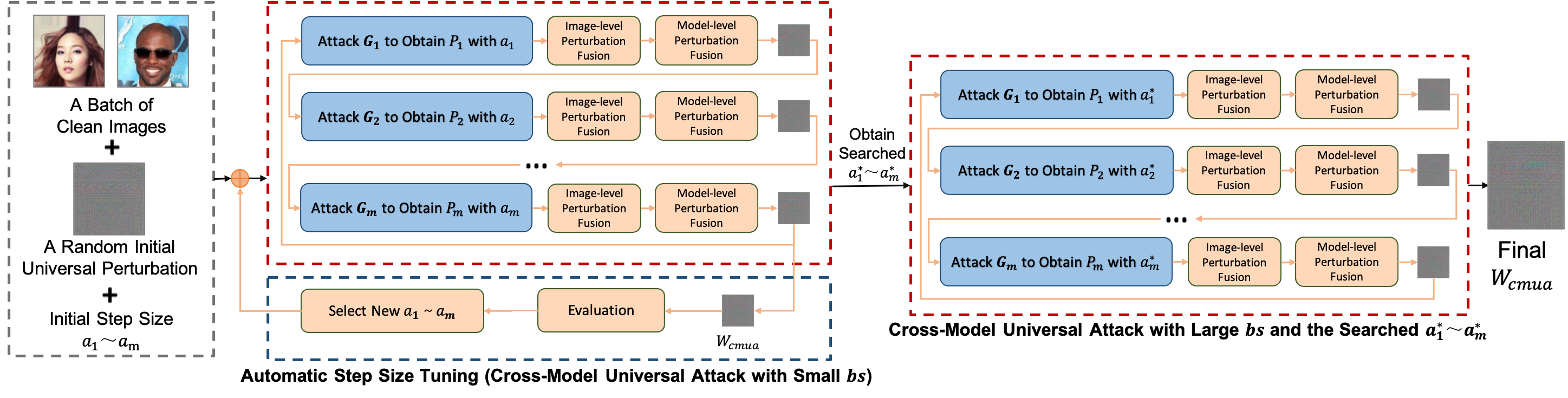}
  \caption{The overall pipeline of our Cross-Model Universal Adversarial Attack on multiple face modification networks.}
  \label{fig:pipeline}
\end{figure*}

\begin{figure}[h]
  \centering
  \includegraphics[width=8.5cm]{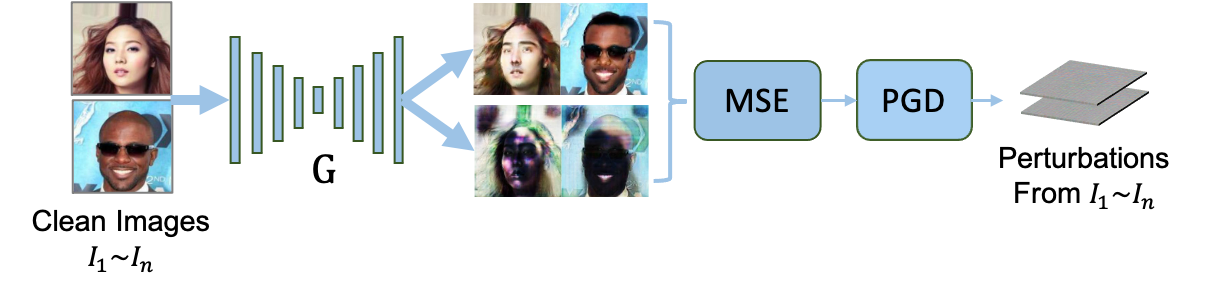}
  \caption{The detailed process of attacking one specific deepfake model.}
  \label{fig:attack_one_model}
\end{figure}

\section{Related Works}

\subsection{Face Modification}

In recent years, the free access to large-scale facial images and the amazing progress of generative models have led face modification networks to generate more realistic facial images with target persons or attributes. StarGAN \cite{choi2018stargan} proposes a novel and scalable approach to perform image-to-image translation across multiple domains, achieving better visual quality in its generated images. Then, AttGAN \cite{AttGAN} uses the attribute-classification constraint to provide more natural facial images on facial attribute manipulation. Moreover, AGGAN \cite{AGGAN} introduces attention masks via a built-in attention mechanism to obtain target images with high quality. Recently, \cite{Li_2021_CVPR} proposes the HiSD which is a state-of-the-art image-to-image translation method for both scalabilities of multiple labels and controllable diversity with impressive disentanglement. Though these models adopt diverse architectures and losses, our CMUA-watermark can successfully prevent facial images from being modified correctly by all of them.

\subsection{Attacks on Generative Models}

There are already some studies \cite{Deceiving-Image-to-Image, Disrupting-Image-Translation-Based, ruiz2020disrupting, kos2018adversarial, tabacof2016adversarial, NEURIPS2019_b83aac23} that explore adversarial attacks on generative models, and we particularly focus on image translation tasks that deepfakes are based upon. \cite{Deceiving-Image-to-Image} and \cite{Disrupting-Image-Translation-Based} attack image translation tasks on CycleGAN \cite{CycleGAN2017} and pix2pixHD \cite{wang2018pix2pixHD}, which only transfer images between two domains and are thus relatively easy to attack. \cite{ruiz2020disrupting} is the first to address attacks on conditional image translation networks, but the watermark they generate only protects a specific image from a specific deepfake model, meaning that every adversarial watermark has to be individually trained, which is time-consuming and infeasible in reality.

\subsection{Universal Adversarial Perturbation}

Universal Adversarial Perturbation is first introduced by \cite{Universal2017}, where a recognition model can be fooled with only a single adversarial perturbation. Here $Universal$ means that a single perturbation can be added to multiple images to fool a specific model. Based on this work, \cite{8237562} introduces universal adversarial perturbation for segmentation tasks to generate target results, and \cite{9010035} first proposes a universal adversarial attack on image retrieval systems, leading them to return irrelevant images. The above-mentioned works only produce a universal adversarial watermark targeting one model, while our CMUA-Watermark can combat multiple face modification models simultaneously.

\begin{table}[]
\caption{The categories of adversarial watermarks.}
\centering
\resizebox{8cm}{!}{
  \label{categories}
\begin{tabular}{@{}ccc@{}}
\toprule
Type           & Cross-Image (i.e., universal) & Cross-Model  \\
\midrule
SIA-Watermark  & \XSolidBrush                     & \XSolidBrush    \\
UA-Watermark & \Checkmark                  & \XSolidBrush    \\
CMUA-Watermark & \Checkmark                  & \Checkmark \\
\bottomrule
\end{tabular}}
\end{table}

\section{Methods}

In this section, we first present an overview of our method. Then, we introduce how to attack a single face modification model. Ensuing, we describe the perturbation fusion strategy. Finally, we analyze the key problems of cross-model optimization and suggest an automatic step size tuning algorithm for searching for suitable attack step sizes.

\begin{algorithm}[htb]
\caption{The Cross-Model Universal Attack} 
\label{alg1}
\begin{algorithmic}[1]

\REQUIRE ~~
 $X_1,...,X_o$ ($o$ batches of training facial images), $bs$ (the batch size), $G_1,...,G_m$ (the combated deepfake models), $a_1,...,a_m$ (the step size of the attack algorithm for $G_1,...,G_m$), $A$ (base attack method which returns the image-and-model-specific adversarial perturbations).

\ENSURE ~~
CMUA-Watermark $W_{cmua}$

\STATE Random Init $W_0$
\FOR{ $k \in [1, o]$ }
    \FOR{ $i \in [1, m]$ }
        \STATE $P_{k_1}^i,...,P_{k_{bs}}^i$ $\Leftarrow$ $A(G_i, a_i, X_k, W_{i+(k-1)m-1})$
        \STATE $P_{avg}^i$ $\Leftarrow$ Image-Level Fusion with $P_{k_1}^i,...,P_{k_{bs}}^i$
        \STATE $W_{i+(k-1)m}$ $\Leftarrow$ Model-Level Fusion with $P_{avg}^i$
    \ENDFOR
\ENDFOR
\STATE $W_{cmua}$ = $W_{m \cdot o}$

\end{algorithmic}
\end{algorithm}

\subsection{Overview}

In Table \ref{categories}, we categorize adversarial watermarks based on their cross-image and cross-model generalization abilities. Differing from Single-Image Adversarial Watermarks (SIA-Watermarks) that protect a specific image against a specific model and Universal Adversarial Watermarks (UA-Watermarks) that protect multiple images against a specific model, the proposed CMUA-Watermark in this paper can combat multiple deepfake models while protecting a myriad of facial images.

As illustrated in Figure \ref{fig:pipeline}, our overall pipeline for crafting the CMUA-Watermark is divided into two steps. In the first step, we repeatedly conduct the cross-model universal attack with a small batch size (for a faster search), evaluate the generated CMUA-Watermark, and then use automatic step size tuning to select new attack step sizes $a_1,...,a_m$. In the second step, we use the found step sizes $a_1^*,...,a_m^*$ to conduct the cross-model universal attack with a large batch size (for enhancing the disrupting capability) and generate the final CMUA-Watermark. Specifically, as shown in Algorithm \ref{alg1}, during the proposed cross-model universal attacking process, batches of input images iteratively go through the PGD \cite{madry2018towards} attack to generate adversarial perturbations, which then go through a two-level perturbation fusion mechanism to combine into a fused CMUA-Watermark that serves as the initial perturbation for the next model.

\subsection{Combating One Face Modification Model}

In this section, we describe the approach to disrupt a single face modification model in our pipeline (the blue box in Figure \ref{fig:pipeline}), illustrated in detail in Figure \ref{fig:attack_one_model}. To begin, we input a batch of clean images $I_1...I_n$ to the deepfake model $G$ and obtain the original outputs $G(I_1)...G(I_n)$. Then, we input $I_1...I_n$ with the initial adversarial perturbation $W$ to $G$, getting the initial distorted outputs $G(I_1+W)...G(I_n+W)$. Subsequently, we utilize Mean Square Error (MSE) to measure the differences between $G(I_1)...G(I_n)$ and $G(I_1+W)...G(I_n+W)$,
\begin{equation}
\max _{W} \sum_{i=1}^{n} MSE(G(I_i), G(I_i+W)), \ s.t. \ \|W\|_{\infty} \leq \epsilon,
\end{equation}
where $\epsilon$ is the upper bound magnitude of the adversarial watermark $W$. Finally, we use PGD \cite{madry2018towards} as the base attack method to update the adversarial perturbations at every attack iteration,
    \begin{equation}
        \begin{split}
         I_{adv}^{0} & = I + W, \\
         I_{adv}^{r+1} = clip_{I, \epsilon}\{I_{adv}^{r} + & a \ sign (\nabla_{I} L(G(I_{adv}^{r}), G(I)))\},\\
        \end{split}
    \end{equation}
where $I$ is the clean facial images, $I_{adv}^{r}$ is the adversarial facial images in the $r$th iteration, $a$ is the step size of the base attack, $L$ is the loss function (we select MSE as formulated in Eq.(2)), $G$ is the face modification network we attack, and the operation $clip$ limits the $I_{adv}$ in the range $[I-\epsilon, I+\epsilon]$. 

Through this process, we can obtain Single-Image-Adversarial Watermarks (SIA-Watermarks) that protect one facial image from one specific deepfake model. However, the crafted SIA-Watermarks are inadequate under the cross-model setting; they are lacking in image-and-model-level transferability. In the following two sections, we introduce our solutions to address this problem.

\subsection{Adversarial Perturbation Fusion}

The conflict among adversarial watermarks generated from different images and models will decrease the transferability of the proposed CMUA-Watermark. To weaken this conflict, we propose a two-level perturbation fusion strategy during the attack process. Specifically, when we attack one specific deepfake model, we conduct an \textbf{image-level} fusion to average the signed gradients from a batch of facial images,
\begin{equation}
    \begin{split}
    & G_{avg} = \frac{\sum_j^{bs} sign (\nabla_{I_j} L(G(I_{j}^{adv}), G(I_j)))}{bs}, 
    \end{split}
\end{equation}
where $bs$ is the batch size of facial images, and $I_{j}^{adv}$ is the $j$th adversarial image of a batch. This operation will cause the $G_{avg}$ to concentrate more on the common attributes of human faces rather than a specific face's attributes. Then, we use PGD to generate the adversarial perturbation $P_{avg}$ though $G_{avg}$ as eq.(2).

After obtaining the $P_{avg}$ from one model, we conduct a \textbf{model-level} fusion, which iteratively combines $P_{avg}$ generated from specific models to the $W_{CMUA}$ in training, and the initial $W_{CMUA}$ is just the $P_{avg}$ calculated from the first deepfake model,
\begin{equation}
    \begin{split}
    & W_{CMUA}^{0} = P_{avg}^{0}, \\
    & W_{CMUA}^{t+1} = \alpha \cdot W_{CMUA}^{t} + (1-\alpha) \cdot \ P_{avg}^{t},
    \end{split}
\end{equation}
where $\alpha$ is a decay factor, $P_{avg}^{t}$ is the averaged perturbation generated from the $t$th combated deepfake model, and $W_{CMUA}^{t}$ is the training CMUA-Watermark after the $t$th attacking deepfake model.

\subsection{Automatic Step Size Tuning based on TPE}

Besides the above mentioned two-level fusion, we find that the attack step sizes for different models are also important for the transferability of the generated CMUA-Watermark. Hence, we exploit a heuristic approach to automatically find the suitable attack step sizes.

The base attack method we select (PGD) falls into the FGSM \cite{GoodfellowSS14} family, and the gradients $\nabla_{x} L$ are normalized by the $sign$ function:
\begin{equation}
    \operatorname{sign} x=\left\{\begin{array}{cll}
        -1 & , & x<0, \\
        0 & , & x=0, \\
        1 & , & x>0.
        \end{array}\right.
\end{equation}
In real calculations, the elements in $\nabla_{x} L$ almost never reach 0, so $||sign(\nabla_{x} L)||_2 \approx 1$ is fixed for any gradient. The updated perturbation $\Delta P$ in an iteration of a sign-based attack method is formulated as:
\begin{equation}
    \Delta P = a \cdot \operatorname{sign}(\nabla_{X} L).
\end{equation}
That is to say, only the step size $a$ determines the update rate during the attack, so the selection of $a$ has a great influence on the attack performance. This conclusion is also valid for cross-model universal attacks; the updated perturbation $\Delta P^u$ in an iteration of a cross-model universal attack is formed by combining $\Delta P_i$ from multiple models $G_1,...,G_m$:
    \begin{equation}
        \Delta P^u = \sum_{i=1}^{m} \alpha^{(m-i)} \Delta P_{i} = \sum_{i}^{m} \alpha^{(m-i)} a_{i} \cdot \operatorname{sign} (\nabla_{X} L_i).
    \end{equation}
In the formula above, $m$ is the number of models, decay factor $\alpha$ is a constant, and $\operatorname{sign} (\nabla_{X} L_i)$ provides an optimization direction for $G_i$. \textbf{Hence, the overall optimization direction is greatly influenced by $a_1,...,a_m$, and selecting the appropriate $a_1,...,a_m$ across different models to find an ideal overall direction is a key problem for cross-model attacks.}

We introduce the TPE \cite{NIPS2011_86e8f7ab} algorithm to solve this problem, automatically searching for the suitable $a_1,...,a_m$ to balance the different directions calculated from multiple models. TPE is a hyper-parameter optimization method based on Sequential Model-Based Optimization (SMBO), which sequentially constructs models to approximate the performance of hyperparameters based on historical measurements, and then subsequently chooses new hyperparameters to test based on this model. In our task, we regard step sizes $a_1,...,a_m$ as the input hyperparameters $x$ and the success rate of the attack as the associated quality score $y$ of TPE. The TPE uses $P(x|y)$ and $P(y)$ to model $P(y|x)$, and $p(x|y)$ is given by:
\begin{equation}
p(x \mid y)=\left\{\begin{array}{ll}
\ell(x), & \text { if } y<y^{*}, \\
g(x), & \text { if } y \geq y^{*},
\end{array}\right.
\end{equation}
where $y^{*}$ is determined by the historically best observation, $\ell(x)$ is the density formed with the observations $\{x^{(i)}\}$ such that the corresponding loss is lower than $y^*$, and $g(x)$ is the density formed with the remaining observations. After modeling the $P(y|x)$, we constantly look for better step sizes by optimizing the Expected Improvement (EI) criterion in every search iteration, which is given by,
\begin{equation}
    \begin{split}
        E I_{y^{*}}(x)= &\frac{\gamma y^{*} \ell(x)-\ell(x) \int_{-\infty}^{y^{*}} p(y) d y}{\gamma \ell(x)+(1-\gamma) g(x)} \\ 
        & \propto\left(\gamma+\frac{g(x)}{\ell(x)}(1-\gamma)\right)^{-1},
    \end{split}
\end{equation}
where $\gamma=p\left(y<y^{*}\right)$. Compared with other criteria, $EI$ is intuitive and has been proven to have an excellent performance. For more details on TPE, refer to \cite{NIPS2011_86e8f7ab}.

\begin{table}
  \caption{The quantitative results of CMUA-Watermark.}
  \label{tab:table_cross_model}
  \resizebox{8.5cm}{!}{
  \begin{tabular}{ccccccc}
  \toprule
\textbf{Dataset} & \textbf{Model} & \textbf{$L_{mask}^2\uparrow$}    & \textbf{$SR_{mask}\uparrow$} & \textbf{FID$\uparrow$} & \textbf{ACS$\downarrow$}                        & \textbf{TFHC$\downarrow$}                                                  \\
    \midrule
    
\multirow{4}{*}{CelebA} 
& StarGAN & 0.20 & 100.00\% & 201.003 & 0.286 & 66.26\%$\rightarrow$20.61\% \\
& AGGAN & 0.13 & 99.88\% & 50.959 & 0.863 & 65.88\%$\rightarrow$55.52\% \\
& AttGAN & 0.05 & 87.08\% & 65.063 & 0.638 & 55.13\%$\rightarrow$28.05\% \\
& HiSD & 0.11 & 99.87\% & 92.734 & 0.153 & 63.30\%$\rightarrow$3.94\% \\
\hline

\multirow{4}{*}{LFW}  
& StarGAN & 0.20 & 100.00\% & 169.329 & 0.207 & 43.88\%$\rightarrow$8.20\% \\
& AGGAN & 0.13 & 99.99\% & 37.746 & 0.806 & 54.90\%$\rightarrow$46.32\% \\
& AttGAN & 0.06 & 94.07\% & 70.640 & 0.496 & 25.86\%$\rightarrow$16.73\% \\
& HiSD & 0.10 & 98.13\% & 88.145 & 0.314 & 50.68\%$\rightarrow$16.03\% \\
  \hline 
  
 \multirow{4}{*}{Film100}  
& StarGAN & 0.20 & 100.00\% & 259.716 & 0.425 & 61.01\%$\rightarrow$29.09\% \\
& AGGAN & 0.13 & 99.88\% & 129.099 & 0.832 & 60.98\%$\rightarrow$55.69\% \\
& AttGAN & 0.07 & 95.82\% & 177.499 & 0.627 & 34.56\%$\rightarrow$25.83\% \\
& HiSD & 0.11 & 100.00\% & 220.689 & 0.207 & 67.00\%$\rightarrow$14.00\% \\
  \bottomrule 
\end{tabular}}
\end{table}

\begin{table*}
\centering
\caption{Comparisons of $SR_{mask}$ and $\log_{10} FID$ with state-of-the-art attack methods.}
\label{tab:compare}
\resizebox{16cm}{!}{
\begin{tabular}{c|cccc|cccc}
\toprule
\multirow{2}{*}{Method}  &  \multicolumn{4}{|c|}{$SR_{mask}\uparrow$} & \multicolumn{4}{c}{$\log_{10} FID\uparrow$} \\ 
& StarGAN & AGGAN & AttGAN & HiSD  & StarGAN & AGGAN & AttGAN & HiSD \\ 
\midrule
BIM \cite{KurakinGB17a}            & 0.6755  & 0.9975       & 0.2126 & 0.0028 & 1.9047 & 1.6539 &1.2451 & 1.6098 \\
MIM \cite{8579055}           & \textbf{1}       & \textbf{0.9994}       & 0.02   & 0.0438 & \textbf{2.5281} & \textbf{1.8435} & 0.7842 & 1.5205\\
PGD \cite{madry2018towards}           & 0.8448  & 0.9970       & 0.0146 & 0.0010 & 2.0203 & 1.6659 & 0.9403 & 1.6467 \\
\text{DI$^2$-FGSM} \cite{xie2019improving}   & 0.028   & 0.3448       & 0.0074 & 0.0001 & 1.5714 & 1.3084 &1.2036 & 1.4113\\
\text{M-DI$^2$-FGSM} \cite{xie2019improving} & \textbf{1}       & 0.9987       & 0.0032 & 0.0050 & 1.5714 & 1.3084 & 1.2036 & 1.4113  \\
AutoPGD \cite{croce2020reliable}       & 0.8314  & 0.9963       & 0.0002 & 0.0007 & 1.5714 & 1.3084 & 1.2036 & 1.4113 \\
Ours           & \textbf{1}       & 0.9988       & \textbf{0.8708} & \textbf{0.9987} & 2.3032 & 1.7072 & \textbf{1.8133} & \textbf{1.9672} \\
\bottomrule 
\end{tabular}}
\end{table*}

\section{Experiment}

In this section, we will first describe our datasets and implementation details. Following, we introduce our evaluation metrics. Then, we show the experimental results of the proposed CMUA-Watermark. Moreover, we systemically conduct an ablation study. Finally, we show an application of the proposed model watermark in realistic scenes.

\subsection{Datasets and Implementation Details}

In our experiments, we use the CelebA \cite{liu2015faceattributes} test set as the main dataset, which contains 19962 facial images. We use the first 128 images in the set as training images and evaluate our method on all facial images of the CelebA test set and the LFW \cite{LFWTech} dataset to ensure credibility. In addition, we also randomly select 100 facial images from films as additional data (Films100) to verify the effectiveness of the CMUA-Watermark in real scenarios. It is important to point out that we do not use any additional data to train our CMUA-Watermark. 

The face modification networks we select in our experiments are StarGAN \cite{choi2018stargan}, AGGAN \cite{AGGAN}, AttGAN \cite{AttGAN}, and HiSD \cite{Li_2021_CVPR}. StarGAN and AGGAN are both trained on the CelebA dataset for five attributes: black hair, blond hair, brown hair, gender, and age. AttGAN is trained on the CelebA dataset for up to fourteen attributes, which is more complicated compared with the two networks above. We also attack one of the latest face modification networks HiSD, which is also trained on the CelebA dataset and can add a pair of glasses to the target person. 

During the process of searching for the step sizes, the maximum number of iterations is 1k, and the search space of the step size for each model is $[0, 10]$. We first search for the step sizes with $batch size=16$ and then use the searched step sizes to conduct cross model attacks with $batch size=64$.

\subsection{Evaluation Metrics}

Considering the limitations of the existing evaluation method and after rethinking the purpose of combating deepfake models, we design a more reasonable and comprehensive evaluation method, which concentrates on metrics of three aspects. 

\begin{figure}[h]
  \small
  \centering
  \includegraphics[width=6cm]{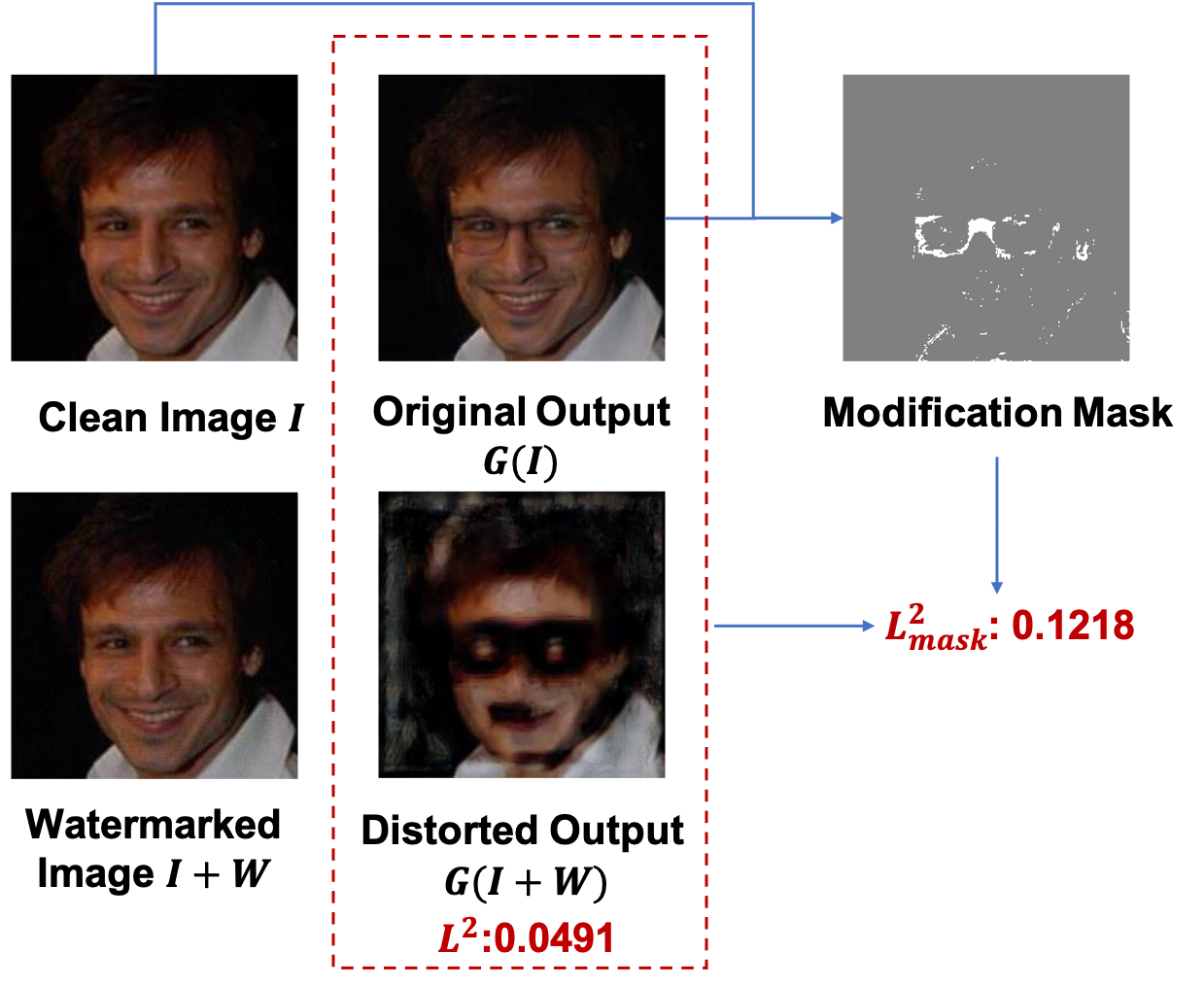}
  \caption{A problematic case for the existing evaluation method and the proposed modification mask for addressing this issue.}
  \label{fig:14}
\end{figure}

To begin, we analyze the success of the modification process. \cite{ruiz2020disrupting} calculates the $L^2$ score between the original outputs $G(I)$ and distorted outputs $G(I+W)$ and declares the facial image as successfully protected by the adversarial watermark when $||G(I+W)-G(I)||_2>0.05$. However, as Figure \ref{fig:14} shows, this evaluation method is problematic for some cases. For example, the distorted output $G(I+W)$ is noticeably different from the original output $G(I)$ in the area of modification, especially around the eyes, hence successfully preventing the deepfake model from adding glasses to the facial image. However, the existing evaluation regards the output as a failed case. To solve this issue, we introduce a mask matrix, concentrating more on the modified areas,
\begin{small}
\begin{equation}
    Mask_{(i,j)} =\left\{\begin{array}{ll}
        1, & if  ||G(I)_{(i,j)} - I_{(i,j)}|| > 0.5, \\
        0, & else,
        \end{array}\right.
\end{equation}
\end{small}
where $(i, j)$ is the coordinate of pixels in the image. In this way, when calculating $L_{mask}^2$, only pixels with major changes will be calculated and other areas will be left out,
\begin{small}
\begin{equation}
    L_{mask}^2 = \frac{\sum_{i} \sum_{j} Mask_{(i,j)} \cdot ||G(I)_{(i,j)} - G(I+W_{CMUA})_{(i,j)}||}{\sum_i \sum_j Mask_{(i,j)}}.
\end{equation}
\end{small}
In our experiments, if the $L_{mask}^2>0.05$, we determine that the image is protected successfully, and use $SR_{mask}$ to represent the success rate of protecting facial images.

Secondly, we use FID \cite{FID} to measure the generation quality of fake facial images. FID comprehensively represents the distance of the feature vectors from Inception v3 \cite{InceptionV3} between original images and fake images, and higher FID values indicate lower qualities of the generated images. 

Finally, we use an open-source liveness detection system HyperFAS\footnote{https://github.com/zeusees/HyperFAS} to test the biological characteristics of the fake images. The HyperFAS is based on Mobilenet \cite{howard2017mobilenets} and trained with 360k facial images. In our experiments, if the confidence score is greater than 0.99, we conclude that the face is a true face with high confidence (\textbf{TFHC}). Additionally, we also calculate the average confidence score \textbf{ACS} for evaluation.

\subsection{The Results of CMUA-Watermark}   
We conduct extensive experiments to demonstrate the effectiveness of the proposed CMUA-Watermark. We first show the quantitative and qualitative results of the proposed CMUA-Watermark and then compare our method with state-of-the-art attack methods. 

The quantitative and qualitative results of the proposed CMUA-Watermark are reported in Table \ref{tab:table_cross_model} and Figure \ref{fig:demo}, respectively. The proposed CMUA-Watermark has a similar overall performance on CelebA and LFW and performs better on StarGAN, AGGAN, and HiSD than on AttGAN. Specifically, the $SR_{mask}$ combating StarGAN, AGGAN, and HiSD is close to 100\% on both CelebA and LFW, and the $ACS$ of the distorted outputs decreases significantly compared with that of the original outputs, which makes the $TFHC$ of StarGAN, AGGAN, AttGAN, and HiSD each drop by 45.65\%, 10.36\%, 27.08\% and 59.36\% on the CelebA test set and 35.68\%, 8.58\%, 9.13\% and 34.65\% on LFW. Besides, on the Film100 dataset that is closer to real scenes, the proposed watermark performs even better compared with the above two datasets. Also, all distorted outputs have large $FID$, demonstrating their poor generation quality. Overall, the above qualitative and quantitative results both demonstrate that the CMUA-Watermark can successfully protect facial images from multiple deepfake models.

We further compare the CMUA-Watermark with state-of-the-art attack methods on CelebA, including BIM \cite{KurakinGB17a}, MIM \cite{8579055}, PGD \cite{madry2018towards}, DI$^2$-FGSM \cite{xie2019improving}, M-DI$^2$-FGSM \cite{xie2019improving}, AutoPGD \cite{croce2020reliable}. We refer to \cite{Universal2017} and adjust these methods to the universal setting (detailed description in the Appendix F). The comparison results of $SR_{mask}$ and $FID$ are reported in Table \ref{tab:compare}, and we can observe that the adversarial watermarks crafted by the compared methods (e.g. MIM) over-optimize on one or two models hence performing very poorly on others. Contrarily, our method achieves excellent performance on all models. These results demonstrate the better image-level and model-level transferability of the proposed method than the existing ones.

\begin{table}[]
\centering
\caption{Ablation study on CelebA test set.}
\label{tab:ablation}
\resizebox{8cm}{!}{
\begin{tabular}{ccccc}
\toprule
\textbf{Methods} & \textbf{PGD}  & &  & \textbf{ours}  \\ \midrule
perturbation fusion       &  & \Checkmark & & \Checkmark  \\
automatic step size tuning         &  &  & \Checkmark & \Checkmark \\ \midrule
$SR_{mask}$(StarGAN)      & 84.5\% & 27\%                      & 99.0\%                      & \textbf{100.0\%} \\
$SR_{mask}$(AGGAN) & 99.7\% & 92\%                      & \textbf{100.0\%}                    & 99.9\%  \\
$SR_{mask}$(AttGAN)       & 1.5\%  & 82.5\%                   & 20.4\%                   & \textbf{87.1\%}  \\
$SR_{mask}$(HiSD)         & 0.1\%  & 0\%                       & 5.0\%                       & \textbf{99.9\%} \\
\bottomrule 
\end{tabular}}
\end{table}

\begin{table}
\centering
\caption{Comparisons of base attack methods on CelebA test set.}
\label{tab:base_attack_compare}
\resizebox{8cm}{!}{
\begin{tabular}{ccccc}
\toprule
\multirow{2}{*}{Base Attack Method}  &  \multicolumn{4}{c}{$SR_{mask}$}      \\   
& StarGAN & AGGAN & AttGAN & HiSD   \\ 
\midrule
CMUA-MIM          & 0.999  & 0.999       & 0.820 & 0.999 \\
CMUA-PGD          & 1.000  & 0.999       & 0.871 & 0.999 \\
\bottomrule 
\end{tabular}}
\end{table}

\subsection{Ablation Study}

In this section, we first investigate the effectiveness of perturbation fusion and automatic step size tuning. Then, we investigate the influence of other hyperparameters on the proposed CMUA-Watermark. As reported in Table \ref{tab:ablation}, the base PGD algorithm has very poor performances on AttGAN and HiSD, indicating that its model-level transferability is weak. When separately using the perturbation fusion strategy the overall performance is improved, but the result for StarGAN drops significantly. On the other hand, if we solely perform automatic step size tuning, the results for all models have been improved but the results for AttGan and HiSD are still not good enough. After combining both of them, the performance of our CMUA-Watermark has improved significantly for all deepfake models. These results demonstrate that perturbation fusion and automatic step size tuning are both crucial to the proposed method and should be used together.

\begin{figure}[h]
  \centering
  \includegraphics[width=7.5cm]{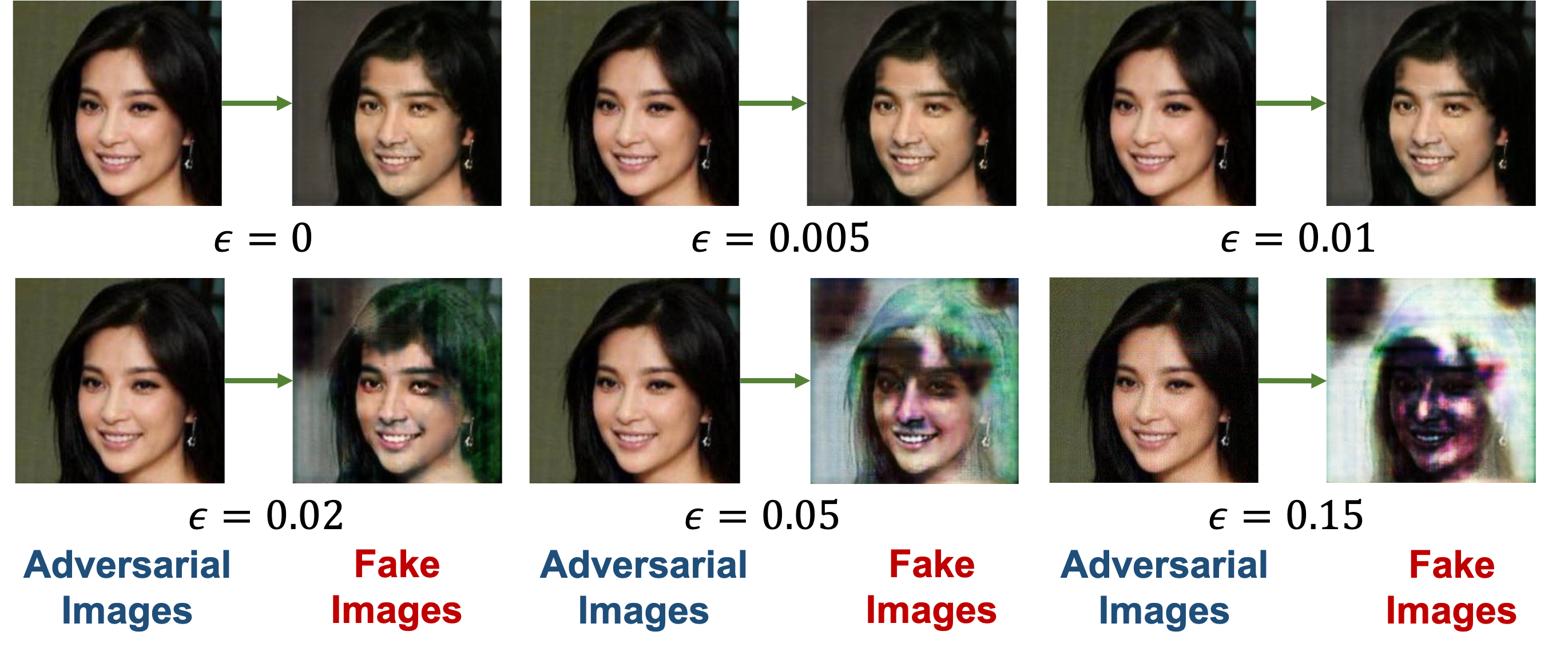}
  \caption{Examples of CMUA-Watermark with different settings of $\epsilon$.}
  \label{fig:epsilon}
\end{figure}

We also investigate the influence on the proposed watermark with the changes of base attack algorithms and the upper bound $\epsilon$. As shown in Table \ref{tab:base_attack_compare}, with the proposed method, changing the base attack method has little influence on the CMUA-Watermark. Besides, as illustrated in Figure \ref{fig:epsilon}, the generated fake facial images are more distorted when the parameter $\epsilon$ becomes larger, meaning that the protection performance becomes better. However, when $\epsilon$ becomes too large, the produced adversarial watermark is more likely to be seen. We empirically find that setting $\epsilon$ around 0.05 can make a good trade-off between the protection performance and the imperceptibility of the produced adversarial watermark.

\begin{figure}[h]
  \centering
  \includegraphics[width=7.5cm]{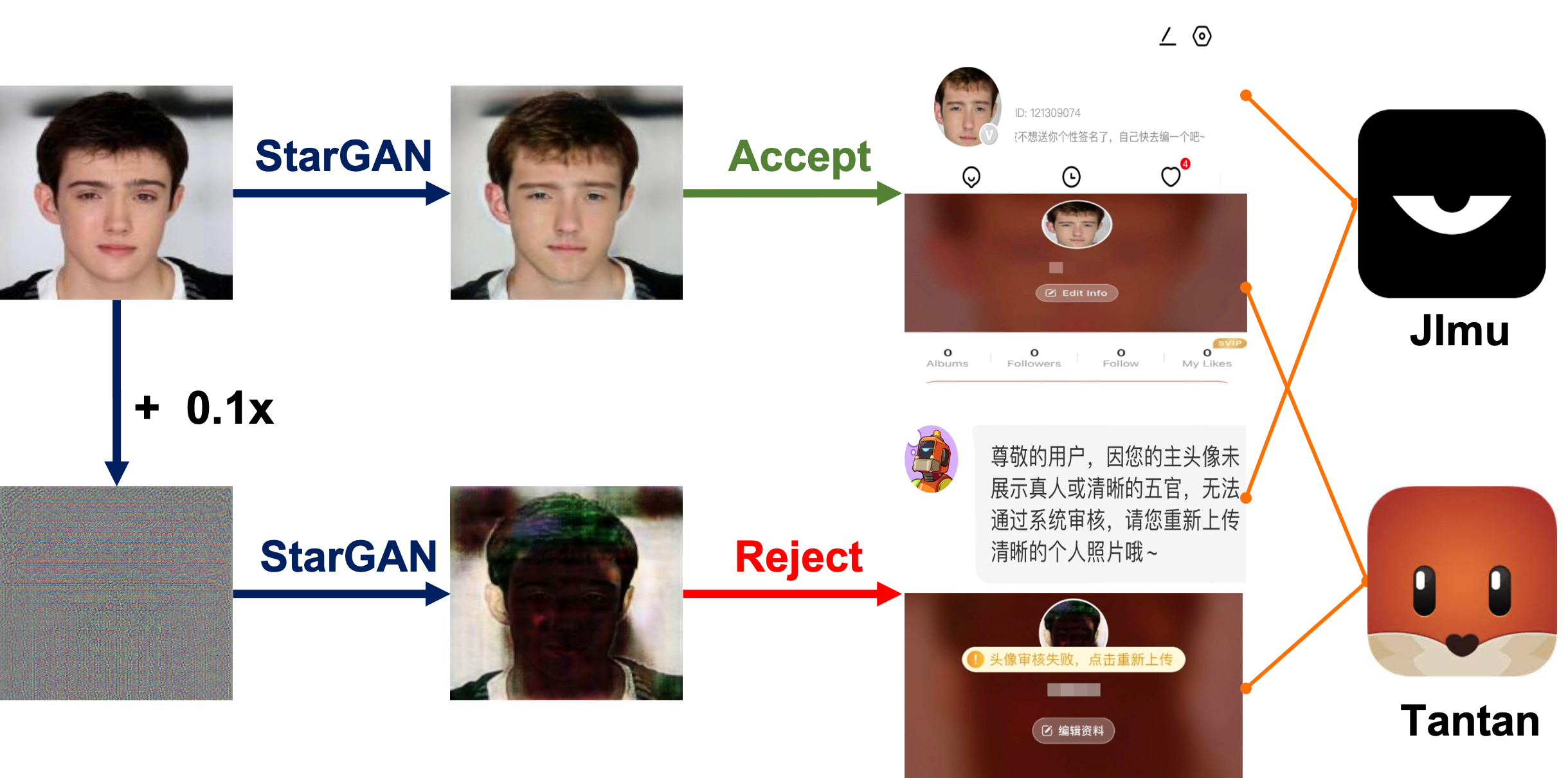}
  \caption{Illustration of CMUA-Watermark preventing fake facial images from passing liveness detection modules integrated in commercial applications.}
  \label{fig:liveness_demo}
\end{figure}

\subsection{Applications}

We further apply the proposed CMUA-Watermark to real social media platforms, which require users to upload their real facial images (e.g., Tantan\footnote{https://apps.apple.com/cn/app/id861891048} and Jimu\footnote{https://apps.apple.com/cn/app/id1094615747}). As illustrated in Figure \ref{fig:liveness_demo}, we find that some fake images generated by StarGAN can pass the verification of liveness detection of Tantan and Jimu successfully. However, the fake images generated from facial images protected by our CMUA-Watermark fail to do so, which demonstrates that our CMUA-Watermark is very effective even in real applications.

\section{Conclusion}

In this paper, we have proposed a cross-model universal attack pipeline to produce a watermark that can protect a large number of facial images from multiple deepfake models. Specifically, we propose a perturbation fusion strategy to alleviate the conflict of adversarial watermarks generated from different images and models in the attack process. Further, we analyze the key problem of cross-model optimization and introduce an automatic step size tuning algorithm based on TPE to determine the overall optimization direction. Moreover, we design a more reasonable and comprehensive evaluation method to evaluate the proposed CMUA-Watermark. Experimental results demonstrate that our CMUA-Watermark can effectively disrupt the modification by deepfake models, lower the quality of generated images, and prevent the fake facial images from passing the verification of liveness detection systems.

\bibliography{aaai22}

\begin{thebibliography}{34}
\providecommand{\natexlab}[1]{#1}

\bibitem[{{Afchar} et~al.(2018){Afchar}, {Nozick}, {Yamagishi}, and
  {Echizen}}]{8630761}
{Afchar}, D.; {Nozick}, V.; {Yamagishi}, J.; and {Echizen}, I. 2018.
\newblock MesoNet: a Compact Facial Video Forgery Detection Network.
\newblock In \emph{2018 IEEE International Workshop on Information Forensics
  and Security (WIFS)}, 1--7.

\bibitem[{Bashkirova, Usman, and Saenko(2019)}]{NEURIPS2019_b83aac23}
Bashkirova, D.; Usman, B.; and Saenko, K. 2019.
\newblock Adversarial Self-Defense for Cycle-Consistent GANs.
\newblock In Wallach, H.; Larochelle, H.; Beygelzimer, A.; d\textquotesingle
  Alch\'{e}-Buc, F.; Fox, E.; and Garnett, R., eds., \emph{Advances in Neural
  Information Processing Systems}, volume~32. Curran Associates, Inc.

\bibitem[{Bergstra et~al.(2011)Bergstra, Bardenet, Bengio, and
  K\'{e}gl}]{NIPS2011_86e8f7ab}
Bergstra, J.; Bardenet, R.; Bengio, Y.; and K\'{e}gl, B. 2011.
\newblock Algorithms for Hyper-Parameter Optimization.
\newblock In Shawe-Taylor, J.; Zemel, R.; Bartlett, P.; Pereira, F.; and
  Weinberger, K.~Q., eds., \emph{Advances in Neural Information Processing
  Systems}, volume~24. Curran Associates, Inc.

\bibitem[{Chen et~al.(2021)Chen, Yao, Chen, Ding, Li, and Ji}]{Chen2021LocalRL}
Chen, S.; Yao, T.; Chen, Y.; Ding, S.; Li, J.; and Ji, R. 2021.
\newblock Local Relation Learning for Face Forgery Detection.
\newblock In \emph{AAAI}.

\bibitem[{Choi et~al.(2018)Choi, Choi, Kim, Ha, Kim, and
  Choo}]{choi2018stargan}
Choi, Y.; Choi, M.; Kim, M.; Ha, J.-W.; Kim, S.; and Choo, J. 2018.
\newblock StarGAN: Unified Generative Adversarial Networks for Multi-Domain
  Image-to-Image Translation.
\newblock In \emph{Proceedings of the IEEE Conference on Computer Vision and
  Pattern Recognition}.

\bibitem[{Croce and Hein(2020)}]{croce2020reliable}
Croce, F.; and Hein, M. 2020.
\newblock Reliable evaluation of adversarial robustness with an ensemble of
  diverse parameter-free attacks.
\newblock In \emph{ICML}.

\bibitem[{Dong et~al.(2018)Dong, Liao, Pang, Su, Zhu, Hu, and Li}]{8579055}
Dong, Y.; Liao, F.; Pang, T.; Su, H.; Zhu, J.; Hu, X.; and Li, J. 2018.
\newblock Boosting Adversarial Attacks with Momentum.
\newblock In \emph{2018 IEEE/CVF Conference on Computer Vision and Pattern
  Recognition (CVPR)}, 9185--9193. Los Alamitos, CA, USA: IEEE Computer
  Society.

\bibitem[{Goodfellow, Shlens, and Szegedy(2015)}]{GoodfellowSS14}
Goodfellow, I.~J.; Shlens, J.; and Szegedy, C. 2015.
\newblock Explaining and Harnessing Adversarial Examples.
\newblock In Bengio, Y.; and LeCun, Y., eds., \emph{3rd International
  Conference on Learning Representations, {ICLR} 2015, San Diego, CA, USA, May
  7-9, 2015, Conference Track Proceedings}.

\bibitem[{{He} et~al.(2019){He}, {Zuo}, {Kan}, {Shan}, and {Chen}}]{AttGAN}
{He}, Z.; {Zuo}, W.; {Kan}, M.; {Shan}, S.; and {Chen}, X. 2019.
\newblock AttGAN: Facial Attribute Editing by Only Changing What You Want.
\newblock \emph{IEEE Transactions on Image Processing}, 28(11): 5464--5478.

\bibitem[{Heusel et~al.(2017)Heusel, Ramsauer, Unterthiner, Nessler, and
  Hochreiter}]{FID}
Heusel, M.; Ramsauer, H.; Unterthiner, T.; Nessler, B.; and Hochreiter, S.
  2017.
\newblock GANs Trained by a Two Time-Scale Update Rule Converge to a Local Nash
  Equilibrium.
\newblock In \emph{Proceedings of the 31st International Conference on Neural
  Information Processing Systems}, NIPS'17, 6629–6640. Red Hook, NY, USA:
  Curran Associates Inc.
\newblock ISBN 9781510860964.

\bibitem[{Howard et~al.(2017)Howard, Zhu, Chen, Kalenichenko, Wang, Weyand,
  Andreetto, and Adam}]{howard2017mobilenets}
Howard, A.~G.; Zhu, M.; Chen, B.; Kalenichenko, D.; Wang, W.; Weyand, T.;
  Andreetto, M.; and Adam, H. 2017.
\newblock MobileNets: Efficient Convolutional Neural Networks for Mobile Vision
  Applications.
\newblock arXiv:1704.04861.

\bibitem[{Huang et~al.(2007)Huang, Ramesh, Berg, and Learned-Miller}]{LFWTech}
Huang, G.~B.; Ramesh, M.; Berg, T.; and Learned-Miller, E. 2007.
\newblock Labeled Faces in the Wild: A Database for Studying Face Recognition
  in Unconstrained Environments.
\newblock Technical Report 07-49, University of Massachusetts, Amherst.

\bibitem[{Korshunov and Marcel(2018)}]{DBLP:journals/corr/abs-1812-08685}
Korshunov, P.; and Marcel, S. 2018.
\newblock DeepFakes: a New Threat to Face Recognition? Assessment and
  Detection.
\newblock \emph{CoRR}, abs/1812.08685.

\bibitem[{Kos, Fischer, and Song(2018)}]{kos2018adversarial}
Kos, J.; Fischer, I.; and Song, D. 2018.
\newblock Adversarial examples for generative models.
\newblock In \emph{2018 ieee security and privacy workshops (spw)}, 36--42.
  IEEE.

\bibitem[{Kurakin, Goodfellow, and Bengio(2017)}]{KurakinGB17a}
Kurakin, A.; Goodfellow, I.~J.; and Bengio, S. 2017.
\newblock Adversarial examples in the physical world.
\newblock In \emph{5th International Conference on Learning Representations,
  {ICLR} 2017, Toulon, France, April 24-26, 2017, Workshop Track Proceedings}.
  OpenReview.net.

\bibitem[{{Li} et~al.(2019){Li}, {Ji}, {Liu}, {Hong}, {Gao}, and
  {Tian}}]{9010035}
{Li}, J.; {Ji}, R.; {Liu}, H.; {Hong}, X.; {Gao}, Y.; and {Tian}, Q. 2019.
\newblock Universal Perturbation Attack Against Image Retrieval.
\newblock In \emph{2019 IEEE/CVF International Conference on Computer Vision
  (ICCV)}, 4898--4907.

\bibitem[{Li et~al.(2021)Li, Zhang, Hu, Cao, Hong, Mao, Huang, Wu, and
  Ji}]{Li_2021_CVPR}
Li, X.; Zhang, S.; Hu, J.; Cao, L.; Hong, X.; Mao, X.; Huang, F.; Wu, Y.; and
  Ji, R. 2021.
\newblock Image-to-Image Translation via Hierarchical Style Disentanglement.
\newblock In \emph{Proceedings of the IEEE/CVF Conference on Computer Vision
  and Pattern Recognition (CVPR)}, 8639--8648.

\bibitem[{Liu et~al.(2015)Liu, Luo, Wang, and Tang}]{liu2015faceattributes}
Liu, Z.; Luo, P.; Wang, X.; and Tang, X. 2015.
\newblock Deep Learning Face Attributes in the Wild.
\newblock In \emph{Proceedings of International Conference on Computer Vision
  (ICCV)}.

\bibitem[{Madry et~al.(2018)Madry, Makelov, Schmidt, Tsipras, and
  Vladu}]{madry2018towards}
Madry, A.; Makelov, A.; Schmidt, L.; Tsipras, D.; and Vladu, A. 2018.
\newblock Towards Deep Learning Models Resistant to Adversarial Attacks.
\newblock In \emph{International Conference on Learning Representations}.

\bibitem[{{Metzen} et~al.(2017){Metzen}, {Kumar}, {Brox}, and
  {Fischer}}]{8237562}
{Metzen}, J.~H.; {Kumar}, M.~C.; {Brox}, T.; and {Fischer}, V. 2017.
\newblock Universal Adversarial Perturbations Against Semantic Image
  Segmentation.
\newblock In \emph{2017 IEEE International Conference on Computer Vision
  (ICCV)}, 2774--2783.

\bibitem[{{Moosavi-Dezfooli} et~al.(2017){Moosavi-Dezfooli}, {Fawzi}, {Fawzi},
  and {Frossard}}]{Universal2017}
{Moosavi-Dezfooli}, S.; {Fawzi}, A.; {Fawzi}, O.; and {Frossard}, P. 2017.
\newblock Universal Adversarial Perturbations.
\newblock In \emph{2017 IEEE Conference on Computer Vision and Pattern
  Recognition (CVPR)}, 86--94.

\bibitem[{Ruiz, Bargal, and Sclaroff(2020)}]{ruiz2020disrupting}
Ruiz, N.; Bargal, S.~A.; and Sclaroff, S. 2020.
\newblock Disrupting deepfakes: Adversarial attacks against conditional image
  translation networks and facial manipulation systems.
\newblock In \emph{European Conference on Computer Vision}, 236--251. Springer.

\bibitem[{Sun et~al.(2021)Sun, Liu, Ye, Gao, Liu, Shao, and
  Ji}]{Sun2021DomainGF}
Sun, K.; Liu, H.; Ye, Q.; Gao, Y.; Liu, J.; Shao, L.; and Ji, R. 2021.
\newblock Domain General Face Forgery Detection by Learning to Weight.
\newblock In \emph{AAAI}.

\bibitem[{Szegedy et~al.(2016)Szegedy, Vanhoucke, Ioffe, Shlens, and
  Wojna}]{InceptionV3}
Szegedy, C.; Vanhoucke, V.; Ioffe, S.; Shlens, J.; and Wojna, Z. 2016.
\newblock Rethinking the Inception Architecture for Computer Vision.

\bibitem[{Tabacof, Tavares, and Valle(2016)}]{tabacof2016adversarial}
Tabacof, P.; Tavares, J.; and Valle, E. 2016.
\newblock Adversarial images for variational autoencoders.
\newblock \emph{arXiv preprint arXiv:1612.00155}.

\bibitem[{Tang et~al.(2019)Tang, Xu, Sebe, and Yan}]{AGGAN}
Tang, H.; Xu, D.; Sebe, N.; and Yan, Y. 2019.
\newblock Attention-Guided Generative Adversarial Networks for Unsupervised
  Image-to-Image Translation.
\newblock In \emph{International Joint Conference on Neural Networks (IJCNN)}.

\bibitem[{Tariq, Lee, and Woo(2021)}]{tariq2021web}
Tariq, S.; Lee, S.; and Woo, S.~S. 2021.
\newblock One Detector to Rule Them All: Towards a General Deepfake Attack
  Detection Framework.
\newblock In \emph{Proceedings of The Web Conference 2021}.

\bibitem[{Tolosana et~al.(2020)Tolosana, Vera-Rodriguez, Fierrez, Morales, and
  Ortega-Garcia}]{TOLOSANA2020131}
Tolosana, R.; Vera-Rodriguez, R.; Fierrez, J.; Morales, A.; and Ortega-Garcia,
  J. 2020.
\newblock Deepfakes and beyond: A Survey of face manipulation and fake
  detection.
\newblock \emph{Information Fusion}, 64: 131--148.

\bibitem[{Wang, Cho, and Yoon(2020)}]{Deceiving-Image-to-Image}
Wang, L.; Cho, W.; and Yoon, K.-J. 2020.
\newblock Deceiving Image-to-Image Translation Networks for Autonomous Driving
  With Adversarial Perturbations.
\newblock \emph{IEEE Robotics and Automation Letters}, PP: 1--1.

\bibitem[{Wang et~al.(2018)Wang, Liu, Zhu, Tao, Kautz, and
  Catanzaro}]{wang2018pix2pixHD}
Wang, T.-C.; Liu, M.-Y.; Zhu, J.-Y.; Tao, A.; Kautz, J.; and Catanzaro, B.
  2018.
\newblock High-Resolution Image Synthesis and Semantic Manipulation with
  Conditional GANs.
\newblock In \emph{Proceedings of the IEEE Conference on Computer Vision and
  Pattern Recognition}.

\bibitem[{Xie et~al.(2019)Xie, Zhang, Zhou, Bai, Wang, Ren, and
  Yuille}]{xie2019improving}
Xie, C.; Zhang, Z.; Zhou, Y.; Bai, S.; Wang, J.; Ren, Z.; and Yuille, A. 2019.
\newblock Improving Transferability of Adversarial Examples with Input
  Diversity.
\newblock In \emph{Computer Vision and Pattern Recognition}. IEEE.

\bibitem[{Yeh et~al.(2020)Yeh, Chen, Tsai, and
  Wang}]{Disrupting-Image-Translation-Based}
Yeh, C.-Y.; Chen, H.-W.; Tsai, S.-L.; and Wang, S.-D. 2020.
\newblock Disrupting Image-Translation-Based DeepFake Algorithms with
  Adversarial Attacks.
\newblock 53--62.

\bibitem[{Zhao et~al.(2021)Zhao, Zhou, Chen, Wei, Zhang, and
  Yu}]{zhao2021multiattentional}
Zhao, H.; Zhou, W.; Chen, D.; Wei, T.; Zhang, W.; and Yu, N. 2021.
\newblock Multi-attentional Deepfake Detection.
\newblock arXiv:2103.02406.

\bibitem[{Zhu et~al.(2017)Zhu, Park, Isola, and Efros}]{CycleGAN2017}
Zhu, J.-Y.; Park, T.; Isola, P.; and Efros, A.~A. 2017.
\newblock Unpaired Image-to-Image Translation using Cycle-Consistent
  Adversarial Networks.
\newblock In \emph{Computer Vision (ICCV), 2017 IEEE International Conference
  on}.

\end{thebibliography}

\end{document}